\documentclass{bmvc2k}
\usepackage{enumitem}
\usepackage{times}
\usepackage{epsfig}
\usepackage{graphicx}
\usepackage{amsmath}
\usepackage{amssymb}

\usepackage{color, colortbl}
\definecolor{LightCyan}{rgb}{0.88,1,1}
\definecolor{Gray}{gray}{0.9}

\usepackage[linesnumbered,ruled,vlined,boxed]{algorithm2e}
\usepackage{multirow}
\usepackage{pifont}

\usepackage{bm}
\SetCommentSty{mycommfont}
\title{Incremental Learning for Animal Pose Estimation using RBF k-DPP}

\addauthor{Gaurav Kumar Nayak*}{gauravnayak@iisc.ac.in}{1}
\addauthor{Het Shah*}{divhet163@gmail.com}{12}
\addauthor{Anirban Chakraborty}{anirban@iisc.ac.in}{1}
\addinstitution{
 Indian Institute of Science\\
 Bangalore, India\\
}
\addinstitution{
 BITS Pilani\\
 Goa, India\\
}

\runninghead{Nayak et al.}{ILAPE using RBF k-DPP}

\def\etal{\emph{et al}\bmvaOneDot}

\begin{document}

\maketitle
\vspace{-0.2in}
\begin{abstract}
Pose estimation is the task of locating keypoints for an object of interest in an image. Animal Pose estimation %deals with predicting poses for different animals and 
is more challenging than estimating human pose due to high inter and intra class variability in animals. % makes which this a tough problem to solve.
Existing works solve this problem for a fixed set of predefined animal categories. Models trained on such sets usually do not work well with new animal categories. Retraining the model on new categories makes the model overfit and leads to catastrophic forgetting. Thus, in this work, we propose a novel problem of ``\textit{Incremental Learning for Animal Pose Estimation}''. Our method uses an exemplar memory, sampled using Determinantal Point Processes (DPP) to continually adapt to new animal categories without forgetting the old ones. We further propose a new variant of k-DPP that uses RBF kernel (termed as ``\textit{RBF k-DPP}'') which gives more gain in performance over traditional k-DPP. Due to memory constraints, the limited number of exemplars along with new class data can lead to class imbalance. We mitigate it by performing image warping as an augmentation technique. This helps in crafting diverse poses, which reduces overfitting and yields further improvement in performance. The efficacy of our proposed approach is demonstrated via extensive experiments and ablations where we obtain significant improvements over state-of-the-art baseline methods.
\end{abstract}

%---------------------------------------------------------------------------------------------------
\vspace{-0.22in}
\section{Introduction}
\label{sec:intro}
\vspace{-0.1in}
Pose estimation is a multi-regression problem that predicts multiple keypoint locations such as Eyes, Nose, Knees, Paws from a given input image (or video). Often the joints are strongly articulated and can be occluded. Thus, pose estimation is quite challenging as it requires searching in the large space of all possible articulated poses. It is an important problem in computer vision due to its utility in several applications. Pose estimation works have inspired human-aware robotics~\cite{svenstrup2009poserobotics,lee2020cameratorobot}, where the robot adapts to human behavior by predicting the human poses. Moreover, there are a lot of behavioral studies on insects and lab animals, such as Drosophila~\cite{gunel2019deepFly3D, pereira2019fastanimalpose, Karashchuk2020anipose} and rats~\cite{Mathis2018DeepLabCut}. Pose estimation has also played a vital role in 3D reconstruction of objects from images/videos~\cite{Loper2015SMPL, kanazawa2018eeRecoveryHumanShapePose, biggs2019SMAL, zuffi2018lionsTigersBears} and action recognition~\cite{luvizon2018poseActionUsingMultitaskDL, AGAHIAN2020efficientHumanActionRecognitionFramework}. 

Human pose estimation, where poses are predicted only for humans, is a fairly popular problem. This is mainly due to large-scale labeled datasets that are publicly available~\cite{andriluka2014MPII, lin2014coco}. A lot of deep learning based pose estimation models have been proposed for this task including OpenPose~\cite{cao2018openpose}, AlphaPose~\cite{fang2018alphapose} and DeepPose~\cite{toshev2014deepPose}. However, Animal pose estimation~\cite{li2020deformation, mackenzie2020dltools, zhang2020omnisupervised}, predicting poses for the animal classes, is comparatively a tougher problem to solve. In general, animals show higher inter and intra class variability compared to humans. %Figure \ref{fig:keypoints_example} shows some instances of animals with their keypoints labels labeled with red dots. 
Due to this variability, training a pose estimation model requires a larger amount of labeled instances. Even labeling such instances is a labor-intensive task that requires basic knowledge about the skeletal composition of the various animal species. Often, due to privacy concerns sharing these datasets publicly may not be possible. Sanakoyeu \etal~\cite{Sanakoyeu2020TransferringDP} publicly released only a subset of the original dataset. To learn animal pose estimation from a limited amount of samples is a challenging problem. \iffalse
\begin{figure}
\begin{tabular}{ccc}
\includegraphics[width=3.66cm]{Figures/cat.png}&
\includegraphics[width=3.66cm]{Figures/cow.png}&
\includegraphics[width=3.66cm]{Figures/horse.png}\\
(a)&(b)&(c)
\end{tabular}
\caption{Examples of some animals with their keypoints labeled. (a) Cat, (b) Cow, (c) Horse}
\label{fig:keypoints_example}
\end{figure}
\fi

In a traditional supervised setting, the set of categories that the pose estimation models can learn to predict poses usually remains fixed after training. However, in a real-world scenario, new categories of animals can keep coming up. Hence, our pose estimation models should also learn to predict poses for these new categories. To achieve this, these traditional systems would require retraining on the newly enhanced training set consisting of new class data and old class data.  
Due to memory constraints, we can only store a subset of old class data. Preserving knowledge for the old classes using only a subset of the dataset is challenging. Hence, selecting samples efficiently in such cases becomes a vital problem. 

Additionally, in the traditional supervised systems, one has to retrain the model from scratch for the new task, which is computationally quite expensive. One may think of using a na\"ive finetuning approach to learn the new categories continually. However, finetuning suffers from overfitting on the new categories causing the pose estimation model to forget the old categories catastrophically. In contrast, humans exhibit the remarkable ability to learn new novel categories and retain previously acquired knowledge. Mimicking this human behavior in artificially intelligent models becomes an interesting and challenging problem at the same time with many exciting applications in the real world. 

Research in the domain of Continual Learning, often referred to as Incremental learning%~\cite{rebuffi2017iCaRL,li2018LwF,castro2018E2EIL,belouadah2019IL2M,zhang2020DMC}
, aims to develop such artificially intelligent systems that can continuously learn new categories while preserving previously acquired knowledge for the old classes. Incremental Learning works are often focused on the problems of Image classification~\cite{rebuffi2017iCaRL, li2018LwF, castro2018E2EIL, belouadah2019IL2M, zhang2020DMC}, Text classification~\cite{chen2007ilTextDocClassification, shan2020learn, sailaja2020iltextCategorization} and Object detection~\cite{shmelkov2017IncrementalLO, brust2020ActiveAI}. However, this problem is still unexplored for Animal Pose estimation. Additionally, the task of pose estimation involving multi-regression is inherently more challenging than classification or object detection.

In this work, we introduce a novel task of ``\textit{Incremental Learning for Animal Pose Estimation}''. In this task, the system is provided with new classes in multiple incremental steps. To mitigate the catastrophic forgetting for old classes, at each incremental step, the system has access to an exemplar memory of fixed size in addition to the new class data. The system has to learn to estimate poses for the new classes without forgetting the old classes. The performance of the incremental learning system should not be affected by the order in which the new classes are added. Thus, we also explore various such scenarios of incrementally learning new classes via several ablation experiments. 

Prior incremental learning works~\cite{castro2018E2EIL, li2018LwF} often add additional heads to the model to learn the new classes. This is memory inefficient, as the number of the parameters grows with the addition of new classes. Our approach does not add any additional layers for incremental steps, keeping the number of parameters for the model constant with the incremental steps. We propose to use Determinantal Point Process (DPP)~\cite{kulesza2010DPP} as a sampling approach to provide a diverse exemplar memory. DPP has been theoretically proven to sample a diverse set of points. To sample a fixed set of k points, we specifically use k-DPP~\cite{kulesza2011kdpp}. We observed that the rank of the kernel is a limitation to the number of points one can sample from k-DPP. The rank of the linear kernel, the kernel used in most of the DPP applications~\cite{kulesza2011kdpp, kulesza2012dppML}, is not sufficient enough to select the required amount of samples. Thus, to overcome this, we propose a new variant of k-DPP that uses RBF Kernel (named as \textit{``RBF k-DPP''}). Our proposed  k-DPP variant overcomes the limitation of the number of points that can be sampled and performs even better than the k-DPP with a linear kernel.
%and performs better than the traditional k-DPP. 
Additionally, to augment the exemplars, we also use an image warping technique to generate varied poses for a given image. 
This helps to overcome the class-imbalance and reduces the overfitting on the new class data, leading to further improvement in performance.

%This augmentation increases the number of samples for old classes preventing the model from overfitting on the new class data and leads to further performance improvement.

We summarize our contribution as follows:
\begin{itemize}[noitemsep,topsep=0pt]
\itemsep0em
\item To the best of our knowledge, we are the first to introduce and solve the novel problem of ``\textit{Incremental Learning for Animal Pose Estimation}''.
\item Our proposed method leverages Determinantal Point Processes (DPPs) %to sample the exemplar memory.
to select samples for the exemplar memory.
\item We propose a new variant of k-DPP that uses RBF Kernel (named as ``RBF k-DPP") which overcomes the sampling limitation of k-DPP. 
\item We also propose to use an image warping-based augmentation technique to generate varied poses for the exemplar memory, which further improves the performance. 
\item The effectiveness of our proposed approach is shown through extensive experiments and improvements over state-of-the-art baselines.
\end{itemize}
%We discuss the proposed definition of ``\textit{Incremental Learning for Animal Pose Estimation}`` and our proposed approach in detail in Sec. % \ref{Chapter3}. 
%We then show our experimental setup in Sec. %\ref{Chapter4} 
%and Results in Sec. % \ref{Chapter5}.

%------------------------------------------------------------------------------------------------------
\vspace{-0.2in}
\section{Related Works}
\label{sec:related_works}
\vspace{-0.1in}
Our work is closely related to Pose estimation, Incremental Learning, DPP, and Image Warping. Thus, we provide a brief description of the relevant works in these domains.

\textbf{Human Pose Estimation:} %There are several Works are abundant in the domain of 
Human pose estimation, both in 2D and 3D, has been extensively studied in the literature. Given that our work focuses mainly on 2D pose estimation, we discuss the works belonging to them. Toshev and Szegedy~\cite{toshev2014deepPose} proposed DeepPose, one of the first major works in human pose estimation using Deep Neural Networks. In recent times, a lot of deep learning based models are proposed such as Stacked Hourglass~\cite{newell2016stackedHourglass}, OpenPose~\cite{cao2018openpose}, MaskRCNN~\cite{he2018maskrcnn}, PoseResnet~\cite{xiao2018poseresnet} and AlphaPose~\cite{fang2018alphapose}. In our current work, we demonstrate our proposed method on the AlphaPose and PoseResnet. However, our framework is independent of the choice of the pose estimation model.

\textbf{Animal Pose Estimation:} 
Few works exist in this domain, and it is relatively less studied than human pose estimation. Some of the recent works have focused on designing pose estimation systems and studying biological functions for specific animals such as Drosophila  flies~\cite{gunel2019deepFly3D, pereira2019fastanimalpose, li2020deformation}, rats~\cite{Mathis2018DeepLabCut} and in the wild animals~\cite{Li2020amur,jiteng2020learningSynAnimals}. There are various toolkits proposed for annotating animal poses, such as DeepPose Kit~\cite{graving2019DeepPoseKit}, DeepLabcut~\cite{Mathis2018DeepLabCut} and AniPose~\cite{Karashchuk2020anipose}. However, these works are limited to specific animal classes. 

Recently, Cao \etal~\cite{cao2019crossDomainAdaptation} and Sanakoyeu \etal~\cite{Sanakoyeu2020TransferringDP} showed an effective domain adaptation strategy %to learn a model 
to predict pose for animal classes using human priors. Li \etal~\cite{Li2021fromSyntheticToReal} and Mu \etal~\cite{jiteng2020learningSynAnimals} proposed unsupervised domain adaptation strategies to predict animal poses. However, our proposed approach does not need any such prior knowledge.
%we, however, do not use human prior knowledge in our work. 
A major limitation of their approach is that it cannot adapt to new animal classes incrementally without retraining the model. 
%If the sampling of exemplars is done incorrectly, the model performance on old classes may deteriorate, eventually leading to low performance on the combined animal classes.
If exemplars are sampled incorrectly, then the model performance on old classes may deteriorate, eventually leading to low performance on the combined animal classes.
%In case proper care in not taken to select exemplars while retraining can lead to overall low performance.
%and would require retraining of the model. 
%On the other hand, our proposed approach can incrementally learn to predict pose for new animal classes and retains the old classes' acquired knowledge.
Our proposed sampling strategy tries to retain the old classes' acquired knowledge.% to solve this problem. 

\textbf{Incremental Learning:} The works in this area are limited to Text Classification~\cite{chen2007ilTextDocClassification, shan2020learn, sailaja2020iltextCategorization}, Object Detection~\cite{shmelkov2017IncrementalLO, brust2020ActiveAI}, and Image Classification~\cite{rebuffi2017iCaRL, li2018LwF, castro2018E2EIL, belouadah2019IL2M, zhang2020DMC}.
One way to mitigate catastrophic forgetting in incremental learning is weight regularization, where a regularization term is added to the loss function~\cite{kirkpatrick2016EWC, aljundi2017MAS}. 
Another way is to use exemplar memory built with samples from the old classes. Sampling a set of exemplars is a challenging task because the small representative set should help the model preserve the knowledge of the old classes. A na\"ive approach is selecting samples randomly from the entire set. This minimizes forgetting to minor extent and is computationally inexpensive~\cite{rebuffi2017iCaRL, chaudry2018rWalk}. GDumb~\cite{prabhu2020gdumb} is another approach that greedily balances the exemplar memory using random sampling. To improve performance further, Herding~\cite{rebuffi2017iCaRL, castro2018E2EIL} is proposed where samples around the class means are chosen as representative samples. Dark Experience Replay (DER)~\cite{buzzega2020detpp} uses reservoir sampling to select the samples for the exemplar memory. 

Many approaches that commonly use exemplar memory further use Knowledge Distillation to regularize the model. 
Rebuffi \etal~\cite{rebuffi2017iCaRL} propose iCaRL that uses an exemplar memory, sampled using herding strategy, and knowledge distillation loss to mitigate catastrophic forgetting. 
iCaRL performs better than LwF~\cite{li2018LwF}, which does not have an exemplar memory and performs distillation on new class data.
Castro \etal~\cite{castro2018E2EIL} improved their performance over iCaRL by adding a balanced finetuning stage to reduce the model's bias towards the new classes. 
However, Belouadah and Popescu~\cite{belouadah2019IL2M} show that even finetuning on exemplars performs better than distillation on them.
Our results show a similar trend. 

\textbf{Image Warping:} This technique is widely used in applications such as surveillance~\cite{Lie1998ApplicationsOfDigitalImageWarping}, Image enhancement~\cite{king1996applicationsForRealTimeImageWarping}. % and video matching~\cite{yuan2017MultiscaleGigapixelVideo}.
Mesh warping using triangular meshes, like Delaunay Triangulation~\cite{lee1980delaunay} is a popular approach to transform images. However, getting mesh labels is a challenging task. A simpler warping technique is Thin-Plate Splines proposed by Bookstein~\cite{bookstein1989TPS}, which takes an image and control points for before and after transformation to transform the grid space. Unlike most works, we use Image warping, Thin-Plate Splines in particular, as an augmentation technique to generate images with diverse poses for the exemplar memory. 

\textbf{DPP:} Selecting samples for exemplar memory is an essential task in incremental learning setup. Determinantal Point Processes (DPPs)~\cite{kulesza2010DPP} are elegant probabilistic models that are theoretically proven to sample a diverse set of points for various tasks, including summarization~\cite{gong2014videosummerizationDPP}, text summarization, image search, and poses~\cite{kulesza2012dppML}. 
A conditional DPP, k-DPP~\cite{kulesza2011kdpp} is often used in real-world applications as it can sample exactly k samples from the DPP. 
%For example, for image search, the goal could be to display ten most relevant images for a search query on Desktop and five for a Mobile Device.
%---------------------------------------------------------------------------------------------------
\vspace{-0.2in}
\section{Proposed Approach}
\label{sec:proposed_approach}
\vspace{-0.1in}
Pose estimation aims to predict the spatial coordinates for a given input image. To train a model in a supervised setting, we have access to the training dataset $\mathcal{D} = (I, G)$, where $I$ are the input images, and $G$ are the ground-truth pose labels.
For an $i^{th}$ image, its ground-truth pose label, $G^i$, can be expressed as,  
% A ground-truth pose  label for an $i^{th}$ image, $G^i$, can be expressed as,
\begin{equation}
    G^i = \{g^i_1, g^i_2, ... g^i_J\} \in {\rm I\!R}^{J \times 3}
\end{equation}
where $J$ denotes the number of keypoints and $g^i_j$ is the $j^{th}$ keypoint label for the $i^{th}$ image. Each keypoint label, $g^i_j$, consists of the spatial coordinates and a value for whether the keypoint is visible in the image or not.
Generally, the keypoint labels for an $i^{th}$ image
are converted to heatmaps, $h^i_j$ for each $j \in \{1, 2, ... J\}$ using the Gaussian Kernel~\cite{Li2020simplepose, fang2018alphapose, cao2018openpose}. Specifically, we center the Gaussian kernel on the spatial coordinates of the keypoint $z^i_j = (x^i_j, y^i_j)$. Mathematically, $h^i_j$ can be formulated as follows, 
\begin{equation}
    h^i_j(x_j, y_j) = \frac{1}{\sqrt{2 \pi \sigma^2}} exp \left( -\frac{[(x-x_j)^2 + (y-y_j)^2]}{2 \sigma^2} \right)
\end{equation}
where $(x,y)$ denote the spatial location in the input image and $\sigma$ is the standard deviation.
The MSE loss between $\boldsymbol{h_{j}}$ and $\boldsymbol{\widehat{h_j}}$ is used for training a supervised pose estimation for all keypoints $j \in \{1,2, ...J\}$, where $\boldsymbol{\widehat{h_j}}$ are the predicted heatmaps.
% \begin{equation}
%     \mathcal{L}_{MSE} = \frac{1}{J} \Sigma_{j=1}^{J} \| \boldsymbol{h_{j}} - \boldsymbol{\widehat{h}_j} \|^2  
% \end{equation}
\begin{figure}
     \centering
     \includegraphics[width=11.0cm]{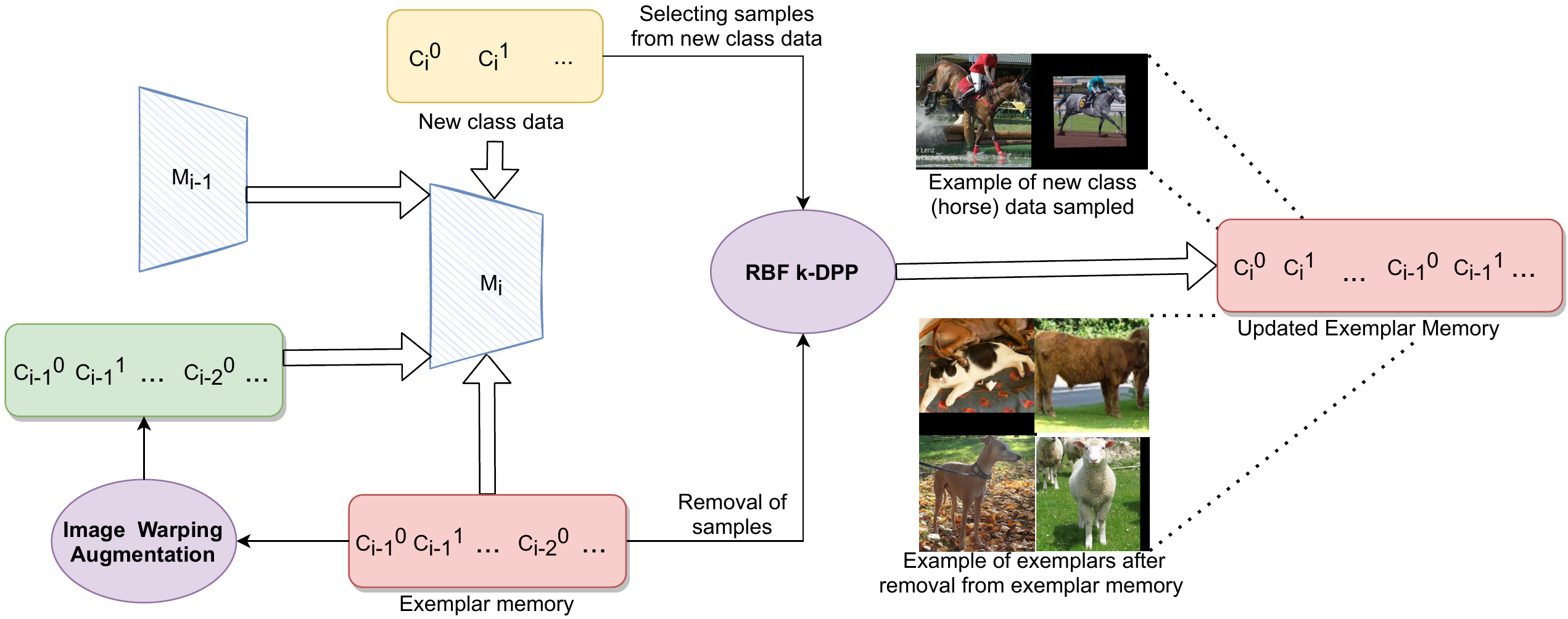}
     \vspace{-0.1in}
     \caption{\small{Our proposed strategy for an $i^{th}$ incremental step is shown. %for Incremental Learning on Animal Pose estimation: 
     %A model at $i^{th}$ incremental step has access to new class data and an Exemplar memory containing the classes that the model has seen at previous incremental step. We further perform augmentation on the exemplar memory to overcome class-imbalance.
     After training the model $M_i$ we update the exemplar memory by selecting new class samples and removing old class samples using RBF k-DPP. 
     %We also perform augmentation on the exemplar memory to overcome class-imbalance.
     Image warping augmentation is further used to overcome class-imbalance.
     }}
     \label{fig:Overview}
\end{figure}
\vspace{-0.3in}

Initially, the pose estimation model, $M_0(\cdot)$, is trained on the dataset $\mathcal{D}_0$  containing ``$a$'' base classes denoted by $C_0 = \{C_0^1, C_0^2, ..., C_0^a\}$ base classes.
At each $i^{th}$ incremental step, we add a dataset $\mathcal{D}_i$ containing ``$b$'' new classes, namely $C_i$,
%= \{C_i^1, C_i^2, ..., C_i^b\}$
where $i \in \{1, 2, ..., k\}$ denotes the incremental step. 
% The dataset for any class $c$ at an incremental step $i$ can be represented as $\mathcal{D}_i^c$.
The goal of the pose estimation model, $M_i(\cdot)$, is to learn to predict keypoints on the new classes, $C_i$, added at the incremental step $i$, without forgetting the old classes, i.e. $C_0 \bigcup C_1 \bigcup C_2 ... \bigcup C_{i-1}$.

To mitigate catastrophic forgetting, we use an exemplar memory~\cite{rebuffi2017iCaRL, castro2018E2EIL, belouadah2019IL2M} consisting of selected samples from old classes data. At each incremental step, the model, $M_i(\cdot)$, has access to the new class data , $\mathcal{D}_{new}$ belonging to the set of classes, $C_i$, and an exemplar memory, $E$, containing data, $\mathcal{D}_{exemplar}$, which consists of the classes that the model has already seen, namely $C_0 \bigcup C_1 \bigcup C_2 ... \bigcup C_{i-1}$. 
% Sampling of the exemplar memory is explained in detail in Section \ref{sec:exemplar_memory}. 
An overview of our proposed approach for Incremental Learning on Animal Pose Estimation is shown in Figure \ref{fig:Overview}.
In the next section, we discuss our method of the selection of samples for the exemplar memory in detail. % is explained in detail Section \ref{sec:exemplar_memory}.
%The performance evaluation of the model, $M_i(\cdot)$ at $i^{th}$ incremental step, is performed on all of the set of classes the model has seen so for, i.e., $C_0 \bigcup C_1 \bigcup C_2 ... \bigcup C_{i}$. 
\vspace{-0.16in}
\subsection{Exemplar memory}
\label{sec:exemplar_memory}
The exemplar memory, $E$, plays an important role in incremental learning to mitigate forgetting. This memory helps the model to preserve the knowledge for the old classes. Thus, we focus on developing the best strategy to select exemplars for %the animal pose estimation 
our pose estimation task that can best represent the classes on which the model is trained. The model uses $E$ as a means to revise the old classes.
We primarily investigate the fixed memory setup, where the memory size is fixed to $\Gamma$. The size of the memory is independent of the number of classes. The number of samples per class, $n$ stored in the exemplar memory, is 
$n = \left\lfloor\frac{\Gamma}{c} \right\rfloor \label{eq:no_samples_memory}$
, where $c$ is the number of classes that the model $M$ has previously seen. After training the model at an incremental step, the exemplar memory is updated by removing samples from the current memory (Sec.~\ref{sec:sampling}) and adding selected samples from the new class data (Sec.~\ref{sec:removal_samples}).
% These operations are discussed further in detail.
% There are two major concerns associated with exemplar memory, which are further discussed in detail. %(1) Sampling the exemplars for the exemplar memory, (2) Removal of samples from the exemplar memory. 
%-------------------------------------------------------------------------------
%  Sampling the exemplar memory 
%-------------------------------------------------------------------------------
\vspace{-0.1in}
% \subsubsection{Sampling the exemplars using DPP}
\subsubsection{Selection of new samples for the exemplar memory using DPP}
\label{sec:sampling}
The exemplar memory plays a pivotal role in the continual learning system, as the model uses only the exemplar memory to retain the knowledge for old classes. A diverse set of exemplars can help the model preserve the acquired knowledge. Selecting samples for the exemplar memory from the new class data is, therefore, a vital task for incremental learning.
%We propose two different approaches to sample the exemplar memory described in the following subsections.
% We propose to use DPP for sampling the exemplar memory which is discussed in the following subsection.
%-------------------------------------------------------------------------------
%  DPP
%-------------------------------------------------------------------------------
% \subsubsection{DPP}

Determinantal point processes (DPP)~\cite{kulesza2010DPP, kulesza2011kdpp, kulesza2012dppML, gong2014videosummerizationDPP} are elegant probabilistic models that can capture negative correlation and are theoretically proven to sample a diverse set of samples. 
Let the set of pose labels for $N$ samples be denoted as $\mathcal{V} = \{G^0, G^1, ... G^N\}$. 
% Then total number of subsets will be the power set of $\mathcal{V}$, $2^{\mathcal{V}}$. 
The probability of selecting a random subset, $S \subseteq \mathcal{V}$ from the power set, $2^{\mathcal{V}}$, is given as: 
\begin{equation}
    P_L(S;L) = \frac{det(L_S)}{det(I+L)}
\end{equation}
where $L$ is real positive semi-definite kernel matrix, referred as L-ensemble \cite{Borodin2002JanossyDI}, 
$L_S = [L_{ij}]_{y_i, y_j \in S}$ is the restriction of $K$ to the rows and columns indexed by $S$, and $I$ is the Identity matrix.
% A point process  $\mathcal{P}$ on a ground set $\mathcal{V}$ is a probabilistic model on the power set of $\mathcal{V}$, $2^{\mathcal{V}}$. The point process $\mathcal{P}$ is called a Determinantal Point Process if for a kernel matrix $K$, which is real and positive semi-definite, and $\textbf{Y}$ is a random subset drawn according to $\mathcal{P}$, then for every $S \subseteq \textbf{Y}$, we have
% \begin{equation}
%     P(S \subseteq \textbf{Y}) = det(K_S)
% \end{equation}
% Here $K_S = [K_{ij}]_{y_i, y_j \in S}$ is the restriction of $K$ to the rows and columns indexed by $S$. In practice, the construction of DPPs is not done through $K$ but via L-ensembles~\cite{Borodin2002JanossyDI}.  An L-ensemble defines a DPP via a real positive semi-definite kernel matrix L and for every $S \subseteq Y$, we have
% \begin{equation}
%     P_L(\textbf{Y} = S) = \frac{det(L_S)}{det(I+L)}
% \end{equation}
% where $I$ is the Identity matrix. 
 
% To find the most likely subset $S$, we can use the MAP estimate that can be formulated as below,
% \begin{equation}
% \label{eq:map}
%     S^{MAP} = argmax_S P_L(\textbf{Y} = S)
% \end{equation}
% Finding this MAP estimate is a NP-Hard problem \cite{ko1995exactAlgoForMaxEntropySampling}. 
However, DPPs have a limitation, they are not constrained on the size of the optimal subset selection. This is not desirable for building exemplar memory in our case, as we need a fixed set of samples from the DPP. Thus, a conditional DPP has to be used. A variation of DPP, k-DPP \cite{kulesza2011kdpp}, models only sets with cardinality $k$. The k-DPP is formulated as,% mathematically observed as follows, 
\begin{equation}
    P_L^k(S;L) = \frac{det(L_S)}{\sum_{|S'|=k} det(L_{S'})}
\end{equation}
where the cardinality of the set S is $k$, i.e. $|S| = k$, and $L$ is a positive semi-definite kernel. It is important to note that $k \leq rank(L)$.
%Similar to the Equation \ref{eq:map}, 
%we can use a MAP estimate to find the most likely set for a k-DPP. 
To find the most likely subset from a k-DPP, we can simply use a MAP estimate, 
\begin{equation}
    S^{MAP}_{k-DPP} = arg max_S P_L^k(S;L)
\end{equation}
Finding the MAP estimate is a NP-Hard problem~\cite{ko1995exactAlgoForMaxEntropySampling}. Thus, it's greedy approximation is used~\cite{kulesza2010DPP, kulesza2012dppML}. %Therefore, a greedy approximation for the MAP estimate is used~\cite{kulesza2010DPP, kulesza2012dppML}.
We discuss two different kernels that are used to construct the k-DPPs.\\
%-------------------------------------------------------------------------------
%  Linear Kernel 
%-------------------------------------------------------------------------------
\textbf{Linear kernel: }
% The linear kernel is most popular kernel used in the literature~\cite{kulesza2012dppML, kulesza2010DPP}. 
The linear kernel~\cite{kulesza2012dppML, kulesza2010DPP} function, $F_L$, is used to calculate the entries, $x_{ij}$ ($i^{th}$ row and $j^{th}$ column), in the L-ensemble kernel matrix. The value of $x_{ij}$ is denoted by $F_L(\boldsymbol{x_i}, \boldsymbol{x_j})$ which is calculated as,
\begin{equation}
    F_L(\boldsymbol{x_i}, \boldsymbol{x_j}) = (\boldsymbol{x_i}\cdot\boldsymbol{x_j})\ \forall \boldsymbol{x_i}, \boldsymbol{x_j} \in \mathcal{V}
\end{equation}
To compute the L-Ensemble matrix ($L$) using $F_L$, we can simply do, $L = A A^T$, where $A$ is the matrix of the pose labels for a particular class.
The shape of this $A$ matrix is $(N, J, 3)$; where $N$ is the number of data points for a particular class and $J$ are the number of keypoints, each keypoint consists of $(x,y)$ co-ordinate location of the keypoint and whether the keypoint is visible or not.  We reshape the $A$ matrix to $(N, J \times 3)$ for simplicity.  Thus, the rank of $L$ is given by 
\begin{equation}
    rank(L) = min(rank(A^T), rank(A))
\end{equation}
The maximum rank of $L$ can thus be $J \times 3$. Thus, the maximum value of $k$, the number of points that can be be sampled in k-DPP is $J \times 3$, which can be very less compared to the required amount of samples. %one may want to sample. 

To overcome the above limitation, we cluster the entire data into $h$ clusters using K-Means%clustering algorithm
~\cite{MacQueen1967SomeMF, franti2019howMuchKMeansbeImproved, david2007KMeans++} based on the pose labels (named as `\textit{k-DPP w/ clustering}'), where $h = \left\lfloor N / (J \times 3) \right\rfloor$.
From each cluster, we sample $J \times 3$ exemplars giving us the desired number of points to be sampled for each class. 
This is computationally expensive because of the computational costs of K-Means and k-DPP.
Due to varying cluster sizes and densities, K-Means can often give sub-optimal clustering of the data. 
% It also depends on the initialization of the cluster centroids, which can heavily influence the performance in selecting a diverse set of points. 
Also, the initialization of cluster centroids can further heavily influence the performance in selecting a diverse set of points. \\
%-------------------------------------------------------------------------------
%  RBF k-DPP
%-------------------------------------------------------------------------------
\textbf{Proposed k-DPP variant (Radial Basis Function (RBF) Kernel based k-DPP): }
To overcome the bottleneck of K-Means clustering for the linear kernel and the computational costs, we use the Radial Basis Function (RBF) to generate the L-ensemble for k-DPP (named as `\textit{RBF k-DPP}'). 
% To use an RBF as an L-Ensemble to construct the k-DPP,
In order to use RBF for constructing an L-Ensemble, 
we need to show that the RBF is positive semi-definite or a kernel. The proof for this can be referred from Supplementary.
The RBF kernel function used to calculate entries, $x_{ij}$ ($i^{th}$ row and $j^{th}$ column), in the L-ensemble kernel matrix is formulated as, 
\begin{equation}
    F_{RBF}(\boldsymbol{x_1}, \boldsymbol{x_2}) = exp(- \gamma\ \| \boldsymbol{x_1} - \boldsymbol{x_2} \|^2)
\end{equation}
where $\gamma \geq 0$ is an hyper-parameter. 

The rank of the RBF Kernel matrix, in the worst case, grows polynomially with the data dimension~\cite{Wang2018numericalRankRBF}. For our case, the rank of the RBF Kernel is sufficiently large enough to sample any amount of samples. This removes the dependency on the use of K-Means clustering 
%does away with the use of K-Means Clustering 
and thus is only required to be applied once per class, thereby reducing the computational costs significantly. Moreover, we observed RBF k-DPP also performs better than the k-DPP w/ clustering.
% Our experimental results further show that the performance of RBF Kernel on the pose estimation task is better than that of the linear kernel with K-Means Clustering.  
\vspace{-0.16in}
\subsubsection{Removal of samples from the Exemplar Memory}
\label{sec:removal_samples}
After each Incremental step, the Exemplar Memory, $E$, is updated to accommodate new classes in the memory. 
To achieve this, we first select a subset of samples for the classes already present in Exemplar memory to make room for new class samples. Removal of samples for the old classes is done by applying the k-DPP again on exemplary memory. %the samples in the memory.
After this operation, the removed samples are never used again. 
% We then sample exemplars from new class and add them to the exemplar memory using the k-DPP. 
%-------------------------------------------------------------------------------
%  Image warping
%-------------------------------------------------------------------------------
\vspace{-0.1in}
\subsection{Image Warping Augmentation}
\label{Augmentation}
To further augment the exemplar memory, we use Thin Plate Splines Image warping (TPS) \cite{bookstein1989TPS} to generate samples with different poses. 
TPS achieves image warping by taking two sets of control points, the input, and the output control points. For our case, we take the limbs visible in the image for an animal, specifically the knee and the paw keypoints.
We rotate these keypoints by a small random angle about the elbow keypoint for the respective limb to get the output control points. The rotation of these keypoints is done under skeletal constraints for each animal. TPS then aligns the input and the output control point sets by warping the grid.  
We further use the image In-painting technique based on the work by Telea~\cite{Telea2004imageInpainting} to fill in the empty portions of the image. 
This in-painting starts from the region's boundary, then gradually %goes inside the region 
fills points towards the center of the region. Each pixel is assigned a normalized value from a small neighborhood of pixels around it. Refer supplementary for visualization of samples, before and after image warping and in-painting operations.
%--------------------------------------------------------------------------------
%  Experiments
%--------------------------------------------------------------------------------
\vspace{-0.11in}
\section{Experiments}
\vspace{-0.09in}
We use Finetuning as our first baseline, where the model is finetuned on the new class data and has no access to the exemplar memory. Further, we compare the performance of our k-DPP constructed using RBF and Linear Kernel against Herding strategy and Random Sampling strategy~\cite{castro2018E2EIL, rebuffi2017iCaRL}. Additionally, we compare our proposed method against some of the heuristic baselines. We modify popular class-incremental learning works on image classification and adapt them to our pose estimation task, which serve as strong baselines for comparison. Oracle denotes the performance of the model when it has access to the entire old class samples, an upper bound for all the approaches.% where the model is trained Finally, we compare against we also use Oracle, where the model is trained on the entire old class dataset, as an upper bound for all the approaches. 
\\
\textbf{Random Sampling Strategy:} This is one of the easiest %and the most intuitive 
ways %of sampling the exemplar memory 
where %We sample 
$n$ samples are randomly selected from the entire set of samples 
%for each class 
to build the memory.\\
\textbf{Herding Strategy:} This has been widely used as a sampling technique for exemplar memory~\cite{castro2018E2EIL, rebuffi2017iCaRL, welling2009herding}. We sort the samples based on the distance from the mean pose for each class. We then select the top-$n$ samples for each class from these sorted lists of samples.% as our exemplar memory.
\\
\textbf{Adapted-iCaRL:} 
We modify iCaRL~\cite{rebuffi2017iCaRL}, originally proposed for classification, %for pose estimation 
to make it suitable for pose estimation task. We replace the cross-entropy loss term with MSE loss for the new class data. We perform distillation on the exemplar memory, built using herding strategy, with MSE loss between the model from the previous incremental step and model at the current step. We give equal weightage to both the MSE loss terms. \\
\textbf{Adapted-EEIL:} Following the work Castro \etal~\cite{castro2018E2EIL}, we add a balanced finetuning step, where equal number of samples for all the classes are considered and distillation loss on the new class data is also applied. \\
\textbf{Adapted-DER++:} We use the Dark Experience Replay (DER) sampling strategy proposed by Buzzega \etal~\cite{buzzega2020detpp} to select the samples for the exemplar memory at each incremental step. We modify the cross-entropy loss with the MSE loss and use the regularization losses proposed in their work. \\ 
\textbf{Adapted-GDumb:} We use the Greedy sampling strategy proposed by Bin \etal~\cite{xiao2018poseresnet} to select the samples for the exemplar memory. We use the MSE loss for training the model at each incremental step.\\
Refer supplementary for more details on adapted baselines.
%----------------------------------------------------------
% Implementation and Setting
%----------------------------------------------------------
\vspace{-0.15in}
\subsection{Implementation and Setting}
We use AlphaPose~\cite{fang2018alphapose} and PoseResnet~\cite{xiao2018poseresnet} as the pose estimation models in our experiments.
We select 17 keypoints similar to Cao \etal~\cite{cao2019crossDomainAdaptation}. 
We set the value of $\sigma = 2$ for generating Gaussian heatmaps.
We perform experiments on the Animal-Pose Dataset \cite{cao2019crossDomainAdaptation} which contains a total of $5117$ instances from five animal classes, namely \{Cat, Cow, Dog, Horse, Sheep\}.
%Total number of samples per class after performing augmentation are \{Cat: 6330, Dog: 7855, Horse: 3800, Sheep: 4390, Cow: 3210\}. }
We consider two different sizes for the fixed exemplar memory, $\Gamma=500$ and $\Gamma=1000$ samples which are only $9.7\%$ and $19.5\%$ of the total dataset size. We repeat the experiments three times and report the mean and standard deviation of the PCK@0.05 values for the predictions made by the pose estimation model on all the classes. 
% Further hyperparameter details can be reffered in the supplementary. 
%---------------------------------------------------
%  Results
%---------------------------------------------------
\vspace{-0.15in}
\subsection{Results and Discussion}
\begin{figure}[htp]
     \centering
     \includegraphics[width=13cm]{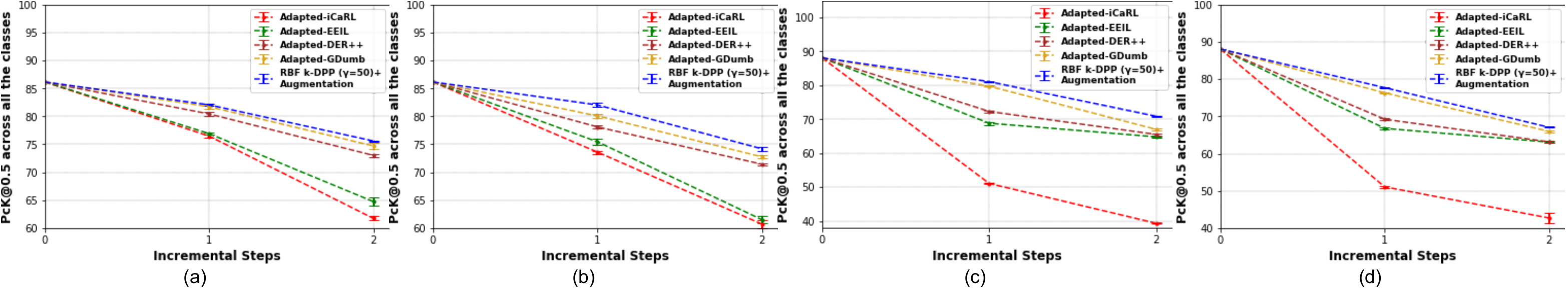}
     \vspace{-0.2in}
     \caption{\small{Experimental Results for different setups of incremental steps. Base classes for (a) and (b) are \{`cat', `cow', `dog'\} while \{`horse'\} and \{`sheep'\} are added at incremental steps. Similarly, the base class for (c) and  (d) is `cat', where \{`horse', `cow'\} and \{`dog', `sheep'\} are added at incremental steps. We set $\Gamma = 1000$ samples for (a) and (c) and $\Gamma = 500$ samples for (b) and (d).}}
     \label{fig:other_incremental}
\end{figure}

We first begin with one base class, `cat', and then incrementally add one animal class at each step in the order of `horse', `cow', `dog', `sheep'.
\begin{table}[htp]
\centering
\scalebox{0.62}{
\setlength{\tabcolsep}{0.12cm}
\begin{tabular}{|c|c|c|c|c|c|c|c|c|}
\hline
\multirow{4}{*}{\textbf{Approach}} & \multicolumn{8}{c|}{\textbf{Memory Sizes ($\Gamma$)}}                                                                                              \\ \cline{2-9} 
                                   & \multicolumn{4}{c|}{\textbf{1000}}                                 & \multicolumn{4}{c|}{\textbf{500}}                                  \\ \cline{2-9} 
                                   & \multicolumn{4}{c|}{\textbf{Incremental steps}}                    & \multicolumn{4}{c|}{\textbf{Incremental steps}}                    \\ \cline{2-9} 
                                   & \textbf{1}   & \textbf{2}   & \textbf{3}   & \textbf{4}            & \textbf{1}   & \textbf{2}   & \textbf{3}   & \textbf{4}            \\ \hline
Oracle                             & 87.49 ± 0.20 & 87.51 ± 0.16 & 84.01 ± 0.22 & 83.47 ± 0.23          & 87.49 ± 0.20 & 87.51 ± 0.16 & 84.01 ± 0.22 & 83.47 ± 0.23          \\ \hline \hline
Finetuning                         & 64.51 ± 0.50 & 56.76 ± 0.16 & 56.56 ± 0.37 & 52.47 ± 0.47          & 64.50 ± 0.49 & 56.76 ± 0.16 & 56.56 ± 0.37 & 52.17 ± 0.47          \\ \hline
Random                             & 83.86 ± 0.16 & 78.14 ± 0.25 & 66.02 ± 0.61 & 64.22 ± 0.36          & 80.64 ± 0.17 & 74.12 ± 0.28 & 64.00 ± 0.22 & 60.88 ± 0.26          \\ \hline
Herding                            & 82.93 ± 0.29 & 75.90 ± 0.34 & 65.96 ± 0.34 & 61.69 ± 0.18          & 80.56 ± 0.14 & 71.82 ± 0.31 & 62.64 ± 0.18 & 59.16 ± 0.27          \\ \hline \hline
\rowcolor{gray!10}k-DPP w/ clustering (\textbf{Ours})           & 84.14 ± 0.12 & 79.07 ± 0.29 & 67.75 ± 0.31 & 65.48 ± 0.43          & 81.63 ± 0.65 & 74.32 ± 0.43 & 64.65 ± 0.14 & 61.71 ± 0.10          \\ \hline
\rowcolor{Gray}RBF k-DPP ($\gamma=0.5$) (\textbf{Ours})         & 84.20 ± 0.48 & 78.99 ± 0.20 & 71.00 ± 0.28 & 68.77 ± 0.31          & 81.56 ± 0.06 & 74.71 ± 0.09 & 64.73 ± 0.12 & 61.88 ± 0.34          \\ \hline
\rowcolor{Gray}RBF k-DPP ($\gamma=50$)  (\textbf{Ours})          & 83.80 ± 0.25 & 79.21 ± 0.35 & 71.38 ± 0.28 & 68.72 ± 0.08          & 81.56 ± 0.30 & 74.59 ± 0.25 & 65.26 ± 0.44 & 62.42 ± 0.27          \\ \hline
\rowcolor{Gray}RBF k-DPP ($\gamma=100$) (\textbf{Ours})          & 83.55 ± 0.11 & 79.21 ± 0.33 & 71.07 ± 0.42 & 68.86 ± 0.52 & 81.54 ± 0.39 & 75.01 ± 0.13 & 65.04 ± 0.18 & 62.46 ± 0.14 \\ \hline
\end{tabular}}
\caption{\small{Mean and Standard deviation of the overall PCK@0.05 of the model across classes are reported in the table. %The same base model is used for all the experiments. 
We show results for two fixed memory sizes, $\Gamma= 500$ and $\Gamma = 1000$ samples. We perform three ablations on $\gamma$ for RBF kernel, with the values 0.5, 50 and 100.
}}
\label{tab:exp1_main_table}
\end{table}
%\vspace{-0.3in}
%\\
\vspace{-0.1in}\\
\textbf{Efficacy of k-DPP:} From Table~\ref{tab:exp1_main_table}, we observe that our proposed approach of k-DPP w/ clustering performs better than the popularly used Herding strategy by $2.5\%$ for the memory $\Gamma=500$ samples, and $3.8\%$ for $\Gamma=1000$ samples, at the final incremental step.  
%Proposed variant RBF k-DPP, yields even better performance. 
For the case of $\Gamma = 500$ samples, we see a small improvement in performance for RBF k-DPP. Note that k-DPP with K-Means has its own shortcomings. It is computationally expensive since we have to perform the sampling using k-DPP for each cluster, whereas in RBF k-DPP, sampling is performed only once. For %the memory of 
$\Gamma = 1000$ samples, we get an improvement of $3.2\%$ in the performance for RBF k-DPP ($\gamma=50$) against k-DPP w/ clustering. The effectiveness of RBF k-DPP can be seen with the increase in the number of incremental steps. Selection of a set of diverse exemplars becomes crucial with the increase in incremental steps as the number of samples per class decreases. RBF k-DPP is more effective in handling this compared to traditional k-DPP with K-Means. 
\begin{table}[htp]
\centering
\scalebox{0.62}{
\setlength{\tabcolsep}{0.13cm}
\begin{tabular}{|c|c|c|c|c|c|c|c|c|}
\hline
\multirow{4}{*}{\textbf{Approach}}                                                 & \multicolumn{8}{c|}{\textbf{Memory sizes ($\Gamma$)}}                                                                                              \\ \cline{2-9} 
                                                                                   & \multicolumn{4}{c|}{\textbf{1000}}                                 & \multicolumn{4}{c|}{\textbf{500}}                                  \\ \cline{2-9} 
                                                                                   & \multicolumn{4}{c|}{\textbf{Incremental steps}}                    & \multicolumn{4}{c|}{\textbf{Incremental Steps}}                    \\ \cline{2-9} 
                                                                                   & \textbf{1}   & \textbf{2}   & \textbf{3}   & \textbf{4}            & \textbf{1}   & \textbf{2}   & \textbf{3}   & \textbf{4}            \\ \hline
k-DPP w/ clustering (\textbf{Ours})                                                               & 84.14 ± 0.12 & 79.07 ± 0.29 & 67.75 ± 0.31 & 65.48 ± 0.43          & 81.63 ± 0.65 & 74.32 ± 0.43 & 64.65 ± 0.14 & 61.71 ± 0.10          \\ \hline 
RBF k-DPP ($\gamma=50$) (\textbf{Ours})                                                          & 83.80 ± 0.25 & 79.21 ± 0.35 & 71.38 ± 0.28 & 68.72 ± 0.08          & 81.56 ± 0.30 & 74.59 ± 0.25 & 65.26 ± 0.44 & 62.42 ± 0.27          \\ \hline \hline
\rowcolor{gray!10}\begin{tabular}[c]{@{}c@{}}RBF k-DPP ($\gamma=50$) +\\ Augmentation (\textbf{Ours})\end{tabular} & 84.57 + 0.22 & 81.88 + 0.35 & 74.52 + 0.44 & \textbf{71.53 + 0.34} & 81.79 ± 0.17 & 74.97 ± 0.06 & 66.49 ± 0.23 & \textbf{63.01 ± 0.11} \\ \hline
\end{tabular}}
\caption{\small{Effect of Augmentation; Performance of our proposed RBF k-DPP ($\gamma$=50) with augmentation against RBF k-DPP($\gamma$=50) and k-DPP w/ clustering.}}
\label{tab:aug_table}
\end{table}
\\
\textbf{Effectiveness of Proposed Augmentation:} Table~\ref{tab:aug_table} shows the effectiveness of our proposed image warping augmentation strategy on top of the proposed DPP variant (RBF k-DPP).  We obtain further improvement of $0.6\%$  and $2.8\%$ at the final incremental step for %the memory sizes of 
$\Gamma = 500$ and $\Gamma = 1000$ samples, respectively. This improvement can be attributed mainly to the enhanced intra-class variance and diversity in poses after augmentation. It thus helps the model to overfit less on the new classes by overcoming class imbalance.
\begin{table}[htp]
\centering
\scalebox{0.64}{
\setlength{\tabcolsep}{0.13cm}
\begin{tabular}{|c|c|c|c|c|c|c|c|c|}
\hline
\multirow{4}{*}{\textbf{Approach}}                                                   & \multicolumn{8}{c|}{\textbf{Memory sizes ($\Gamma$)}}                                                                                              \\ \cline{2-9} 
                                                                                     & \multicolumn{4}{c|}{\textbf{1000}}                                 & \multicolumn{4}{c|}{\textbf{500}}                                  \\ \cline{2-9} 
                                                                                     & \multicolumn{4}{c|}{\textbf{Incremental steps}}                    & \multicolumn{4}{c|}{\textbf{Incremental Steps}}                    \\ \cline{2-9} 
                                                                                     & \textbf{1}   & \textbf{2}   & \textbf{3}   & \textbf{4}            & \textbf{1}   & \textbf{2}   & \textbf{3}   & \textbf{4}            \\ \hline
Adapted - iCaRL                                                                      & 61.90 ± 0.09 & 48.49 ± 0.43 & 38.51 ± 0.39 & 34.88 ± 0.47          & 61.64 ± 0.49 & 47.15 ± 0.50 & 39.24 ± 2.62 & 37.53 ± 1.16          \\ \hline
Adapted - EEIL                                                                       & 72.17 ± 0.63 & 64.86 ± 0.66 & 60.43 ± 0.69 & 57.41 ± 0.01          & 69.15 ± 0.15 & 59.99 ± 0.32 & 58.64 ± 0.31 & 56.05 ± 0.04          \\ \hline
Adapted - DER++                                                & 76.29 ± 0.32 & 68.93 ± 0.71 & 59.60 ± 0.12 & 56.28 ± 0.25          & 72.56 ± 0.33 & 64.80 ± 0.43 & 55.57 ± 0.13 & 51.67 ± 0.52         \\
\hline
Adapted - GDumb                                                  & 83.11 ± 0.16 & 77.48 ± 0.35 & 66.89 ± 0.23 & 63.22 ± 0.42          & 81.12 ± 0.28 & 73.20 ± 0.17 & 63.93 ± 0.22 & 60.40 ± 0.46          \\
\hline \hline
\rowcolor{gray!10}\begin{tabular}[c]{@{}c@{}}RBF k-DPP ($\gamma=50$)  +\\  Augmentation (\textbf{Ours})\end{tabular} & 84.57 + 0.22 & 81.88 + 0.35 & 74.52 + 0.44 & \textbf{71.53 + 0.34} & 81.79 ± 0.17 & 74.97 ± 0.06 & 66.49 ± 0.23 & \textbf{63.01 ± 0.11} \\ \hline

\end{tabular}}
\caption{\small{Performance of our proposed RBF k-DPP ($\gamma$=50) with augmentation against adapted incremental learning baselines using AlphaPose model~\cite{fang2018alphapose}}}
\label{tab:sota_table}
\end{table}
\\
\textbf{Comparison against Adapted Incremental Learning Baselines:} We compare our pose incremental learning model's performance against the state-of-the-art iCaRL, EEIL, DER++ and GDumb adapted for the pose estimation task in Table~\ref{tab:sota_table}. We observe a significant improvement of $2.6\%$, $11.3\%$, $7\%$ and $25.5\%$ in performance against GDumb, DER++, EEIL and iCaRL, respectively at $\Gamma = 500$ samples. Further, at %the memory of 
$\Gamma = 1000$ samples, our proposed method again obtains significant improvement of $8.31\%$, $15.25\%$, $14.1\%$ and $36.6\%$ over GDumb, DER++, EEIL and iCaRL, respectively. \\
\textbf{Other Incremental Learning Setups:} We further perform experiments with different setups for incremental steps as shown in Figure~\ref{fig:other_incremental}. Our proposed approach consistently performs better than the adapted versions of the state-of-the-art works.
\\
\textbf{Comparison with other Augmentation Strategy:} In Table~\ref{tab:aug}, we compare the performance of our proposed Image Warping augmentation method with other augmentation strategies, namely Rotation, Flipping and Gaussian Noise. We obtain $1.5\%$, $0.3\%$ and $0.1\%$ improvement in performance over Noise, Rotation and Flipping respectively at $\Gamma = 1000$. Also, we further observe an improvement of $1.7\%$, $0.9\%$ and $0.8\%$ compared to Noise, Rotation and Flipping respectively for $\Gamma=500$.
\begin{table}[htp]
\centering
\scalebox{0.64}{
\setlength{\tabcolsep}{0.13cm}
\begin{tabular}{|c|c|c|c|c|}
\hline
\multirow{4}{*}{\textbf{Approach}}                                                   & \multicolumn{4}{c|}{\textbf{Memory sizes ($\Gamma$)}}                                                        \\ \cline{2-5} 
                                                                                     & \multicolumn{2}{c|}{\textbf{1000}}              & \multicolumn{2}{c|}{\textbf{500}}               \\ \cline{2-5} 
                                                                                     & \multicolumn{2}{c|}{\textbf{Incremental steps}} & \multicolumn{2}{c|}{\textbf{Incremental steps}} \\ \cline{2-5} 
                                                                                     & \textbf{1}         & \textbf{2}                 & \textbf{1}         & \textbf{2}                 \\ \hline
RBF k-DPP ($\gamma=50$)  + Flipping                                                       & 81.53 ± 0.23     & 75.46 ± 0.35               & 81.53 ± 0.28       & 73.41 ± 0.36               \\ \hline
RBF k-DPP ($\gamma=50$) + Rotation                                                        & 81.84 ± 0.42     & 75.31 ± 0.24               & 81.13 ± 0.24       & 73.29 ± 0.51               \\ \hline
RBF k-DPP ($\gamma=50$) + Noise                                                           & 81.81 ± 0.18     & 74.09 ± 0.19               & 81.39 ± 0.42       & 72.53 ± 0.25               \\ \hline \hline
\rowcolor{gray!10}\begin{tabular}[c]{@{}c@{}}RBF k-DPP ($\gamma=50$)  + \\ Augmentation (\textbf{Ours})\end{tabular} & 82.05 ± 0.15       & \textbf{75.59 ± 0.13}      & 82.03 ± 0.34       & \textbf{74.19 ± 0.33}      \\ \hline
\end{tabular}}
\caption{\small{Experimental results of our proposed augmentation strategy against other augmentation strategies, Flipping, Rotation and Gaussian Noise. We consider \{`cat', `cow', `dog'\} as base classes while \{`horse'\} and \{`sheep'\} are added at incremental steps.}}
\label{tab:aug}
\end{table}

\textbf{Experiments on PoseResnet model~\cite{xiao2018poseresnet}:} We compare the performance of our proposed approach against the adapted state-of-the-art Incremental Learning baselines iCaRL, EEIL, DER++ and GDumb on PoseResnet~\cite{xiao2018poseresnet} architecture. As observed in Table~\ref{tab:pose_resnet}, our proposed approach achieves better performance than the adapted baselines for both the exemplar memory settings ($\approx 4.2\%$ for $\Gamma=1000$ and $\approx 1.5\%$ for $\Gamma=500$).
% Similarly, we perform experiments with different setups for the incremental steps. have $\{ `cat`, `cow`, `dog`\}$ as the base classes and at each incremental step add one class of `horse`  followed by `sheep`. In the final setup, we have `cat` as the base class and we add two classes at each incremental step namely, $\{`horse`, `cow`\}$ followed by $\{`dog`, `sheep`\}$.
\begin{table}[htp]
\centering
\scalebox{0.64}{
\setlength{\tabcolsep}{0.2cm}
\begin{tabular}{|c|c|c|c|c|}
\hline
\multirow{4}{*}{\textbf{Approach}}                                                   & \multicolumn{4}{c|}{\textbf{Memory sizes ($\Gamma$)}}                                                        \\ \cline{2-5}
                                                                                     & \multicolumn{2}{c|}{\textbf{1000}}              & \multicolumn{2}{c|}{\textbf{500}}               \\ \cline{2-5}
                                                                                     & \multicolumn{2}{c|}{\textbf{Incremental steps}} & \multicolumn{2}{c|}{\textbf{Incremental Steps}} \\ \cline{2-5}
                                                                                     & \textbf{1}             & \textbf{2}            & \textbf{1}             & \textbf{2}              \\ \hline
Adapted - iCaRL                                                                      & 50.30 ± 0.31           & 37.80 ± 0.40          & 46.60 ± 0.26           & 35.93 ± 0.26                  \\ \hline
Adapted - EEIL                                                                       & 45.30 ± 0.19           & 39.97 ± 0.46          & 41.33 ± 0.36           & 38.72 ± 0.55                  \\ \hline
Adapted - DER++                                                                      & 54.50 ± 0.24           & 47.92 ± 0.31          & 52.00 ± 0.30           & 47.27 ± 0.35                 \\ \hline
Adapted - GDumb                                                                      & 55.35 ± 0.28           & 51.45 ± 0.49          & 51.21 ± 0.43           & 47.64 ± 0.39                  \\ \hline \hline
\rowcolor{gray!10}\begin{tabular}[c]{@{}c@{}}RBF-DPP ($\gamma=50$)  + \\ Augmentation (\textbf{Ours})\end{tabular} & 57.93 ± 0.32              & \textbf{55.72 ± 0.22}        & 52.45 ± 0.31                  & \textbf{49.1 ± 0.36}  \\ \hline       
\end{tabular}}
\caption{\small{Performance of our proposed RBF k-DPP ($\gamma=50$) against adapted incremental learning baselines using PoseResnet model~\cite{xiao2018poseresnet}. We consider \{`cat', `cow', `dog'\} as base classes while \{`horse'\} and \{`sheep'\} are added at incremental steps.}}
\label{tab:pose_resnet}
\end{table}

%--------------------------------------------------------------
%  Conclusion
%--------------------------------------------------------------
\vspace{-0.27in}
\section{Conclusion}
This paper proposes a novel task of %`\emph{Incremental Learning for Animal Pose Estimation}'' 
incrementally learning to predict pose for animals. Our experimental results %on Animal Pose Dataset 
show:  
i) selection of samples for the exemplar memory %from the new class data 
plays a pivotal role in incremental learning, 
%i) sampling an efficient exemplar memory plays a pivotal role in incremental learning, 
ii) the effectiveness of proposed k-DPP over the popular heuristic baselines,
iii) our proposed RBF k-DPP performs significantly better than state-of-the-art baselines and yields more gain over k-DPP besides being less computationally expensive, 
iv) augmentation using image warping handles class imbalance and improves performance.

\iffalse
\section{Acknowledgements}
This work is supported by a project grant from Wipro IISc Research and Innovation Network (WIRIN). We would like to extend our gratitude to all the reviewers for their valuable suggestions. We would also like to thank Siddharth Seth for helpful discussions on this project.
\vspace{-0.2in}
\fi

\newpage
\onecolumn
\begin{center}
    \LARGE{\textbf{ \textit{Supplementary for\\} ``Incremental Learning for Animal Pose Estimation using RBF k-DPP''} }

\end{center}

\setcounter{section}{0}
\setcounter{table}{0}
\setcounter{figure}{0}
\setcounter{equation}{0}
\vspace{8pt}
\hrule
\vspace{18pt}

\section{Detailed Diagrammatic View of our Approach}
%\vspace{-0.15in}
%-----------------------------------------------------
\begin{figure}[htp]
     \centering
     \includegraphics[width=\textwidth]{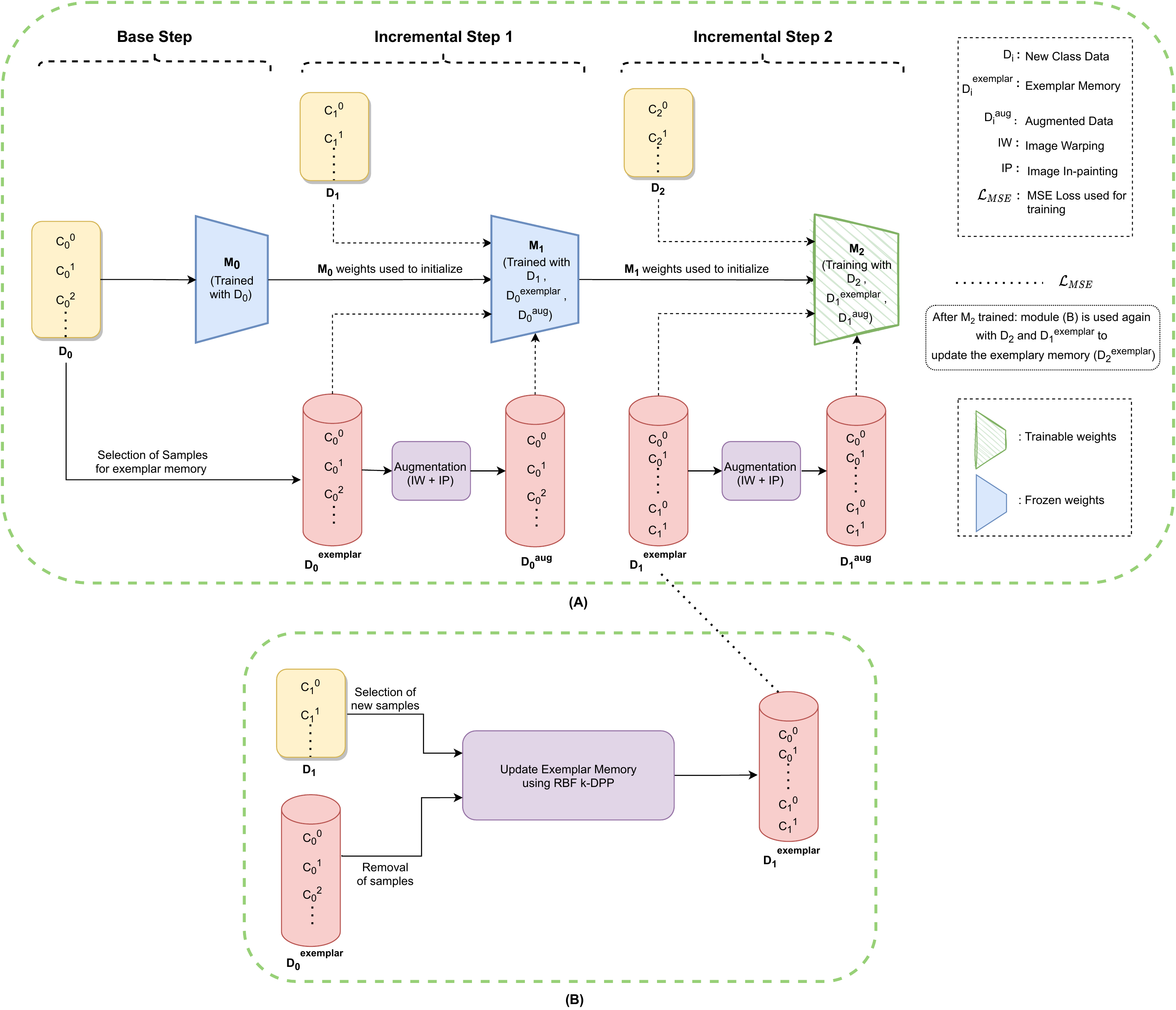}
     \small
     \caption{\small{(A) Complete overview of our approach. We depict training process at the second incremental step. 
     The model $\boldsymbol{M_2}$ is initialized with the weights of $\boldsymbol{M_1}$ and its weights are kept trainable. % is trained using $D_2$, $D_1^{exemplar}$ and $D_1^{aug}$. 
     Augmentation module is explained in detail in Figure \ref{fig:warping_overview}. (B) To update the exemplar memory, we remove samples from $D_0^{exemplar}$, and select new samples from new class data, $D_1$. Thus, $D_1^{exemplar}$ is obtained after this update step which is used along with $D_1^{aug}$ and $D_2$ to train $\boldsymbol{M_2}$.}
     }
     \label{fig:IL_overview}
\end{figure}
%\pagebreak
%\vspace{-0.2in}
\newpage
\section{Image Warping Augmentation}
\begin{figure}[htp]
     \centering
     \includegraphics[width=\textwidth]{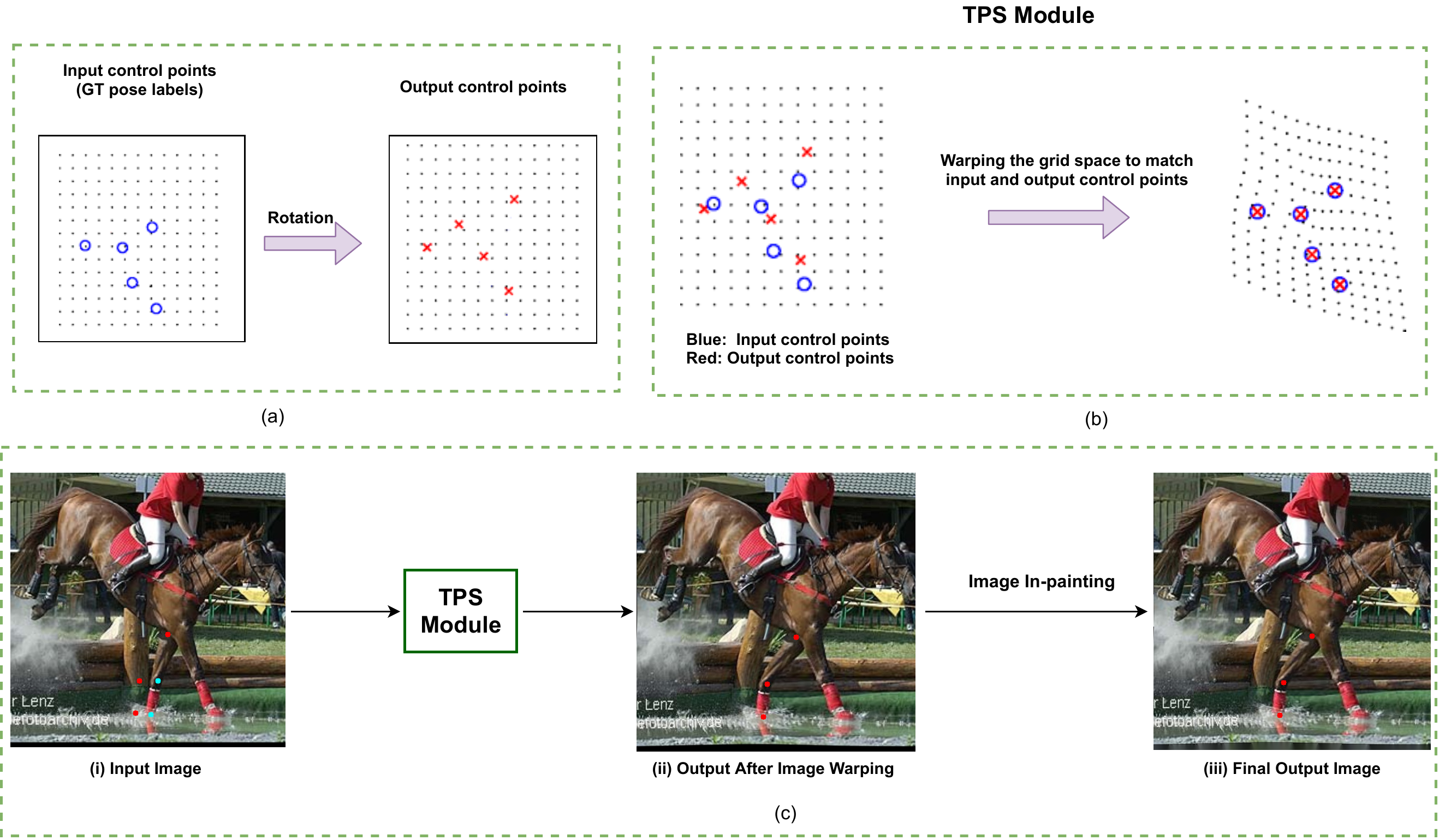}
     \caption{Overview of Image warping and in-painting augmentation. (a) Ground truth pose labels are taken as Input control points, we rotate these keypoints by a small angle under skeletal constraints for each animal to get the Output control points, (b) TPS warps the image grid space to match the input and output control points, (c) We show steps of the proposed augmentation technique (i) For demonstration we only rotate the left frontal leg of the horse, cyan dots represent the original keypoints and red dots represent the rotated keypoints, (ii) After TPS warping, we get the rotated left front limb, (iii) Final output image is generated after applying image in-painting.}
     \label{fig:warping_overview}
\end{figure}

Figure \ref{fig:warping_overview}, describes the image warping and in-painting operations which are used as an augmentation technique and are applied on the input images of animal pose data. The limbs of horse in Figure \ref{fig:warping_overview} (c) are rotated by a small angle using Thin Plate Splines (TPS Module). We then perform image in-painting to fill in the pixels with no values. 

\begin{figure}[htp]
    \centering
    \includegraphics[width=0.8\textwidth]{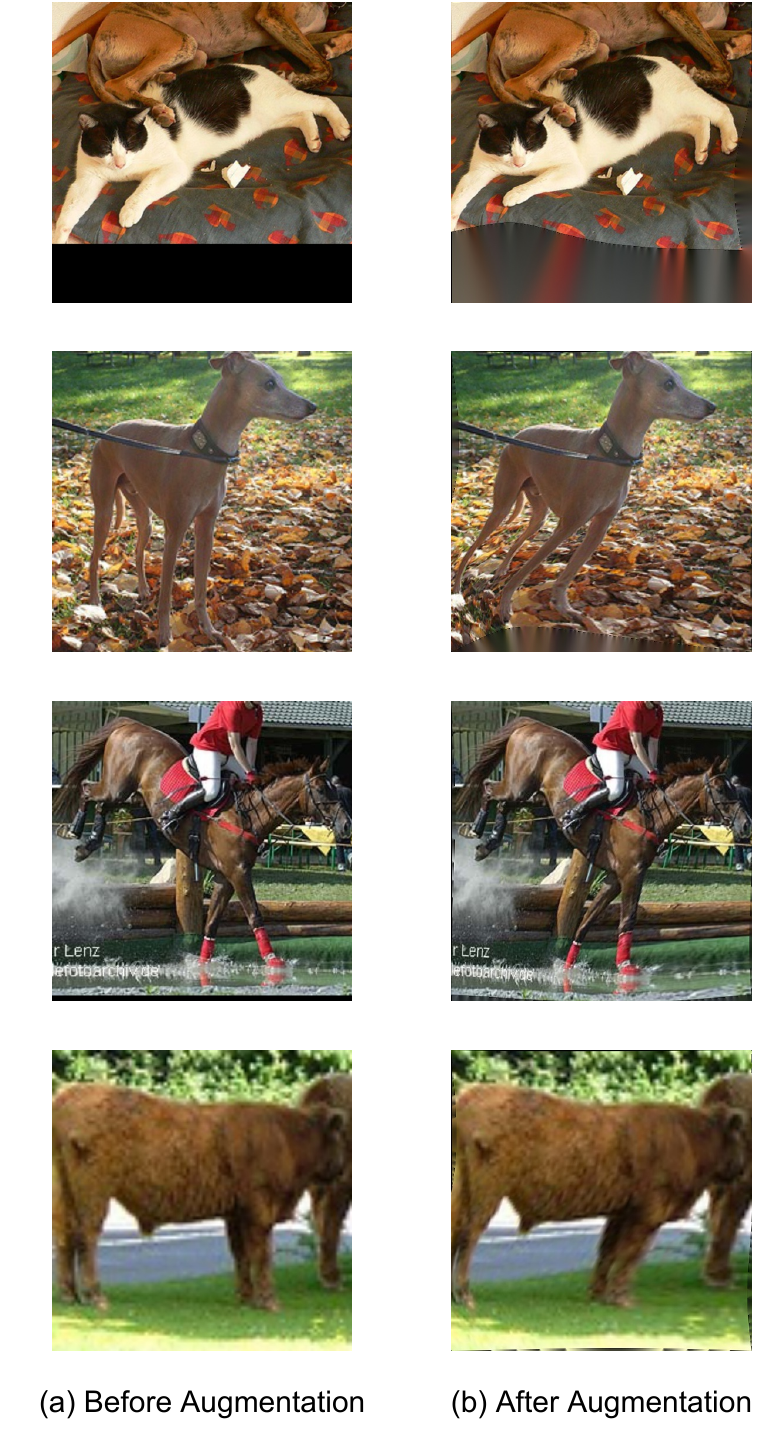}
    \caption{The images in (a) are before augmentation, images in (b) are after image warping and in-painting augmentation applied to respective images in (a).}
    \label{fig:augmentation}
\end{figure}

The proposed augmentation when applied to samples from the exemplar memory helps in creating diverse poses. This eventually mitigates class imbalance between new class data and exemplar memory during the training at each incremental step. Figure \ref{fig:augmentation} demonstrates samples generated using this augmentation technique. 
%\newpage
\newpage
\section{Proof for ``RBF is positive semi-definite''}
The Radial Basis Function, $F_{RBF}(x, y)$, when x, y are real numbers, is given as
\begin{equation}
    F_{RBF} (x, y) = exp(-\frac{\| x - y \|^2}{2\sigma^2})
\end{equation} 
% is based on the Bochner's Theorem \cite{Helson1986BochnerTheorem} 
% and is given as follows - 
Without the loss of generality, we can assume that $\sigma = 1$.
We can further, write $F_{RBF}(x, y) = h(x-y)$, where 
\begin{equation}
\label{eq:moment}
\begin{split}
    h(t) 
    & = exp(-\frac{t^2}{2}) \\
    & = exp(0(it) + 1/2(1)^2(it)^2), 
\end{split}
\end{equation}
here $i$ denotes iota.

Let's assume that Z is a random variable where 
$Z \sim G(0,1)$, G is the Gaussian distribution. 
% follows Gaussian distribution with mean=0 and  standard deviation=1, 
and we know that moment generating function for Gaussian distribution can be written as, 
\begin{equation}
\label{eq:moment_gaussian}
    M_Z(x) = exp(\mu x + (\sigma^2 x^2)/2)
\end{equation}
From Eq. \ref{eq:moment} and Eq. \ref{eq:moment_gaussian} we get, 
\begin{equation}
    \begin{split}
    h(t)
    & = M_Z \left( it \right) \\
    & = E[e^{itZ}]
    \end{split}
\end{equation}
% Further a function is positive semi-definite if $Q\geq0$, where Q is a quadratic form of the function.
Any $m \times m$ matrix $A$ is positive semi-definite if, 
\begin{equation}
    \boldsymbol{v}^T A \boldsymbol{v} \geq 0, \hspace{0.1in} \forall \textbf{v} \in \mathbb{R}^m
\end{equation}
% Further we know that if for any quadratic form Q of a function, $Q\geq0$ then the function is positive semi-definite. 
Thus, for real numbers $x_1, x_2, ... x_n$ and $a_1, a_2, ... a_n$, a quadratic form of $F_{RBF}$, would be  
%it is sufficient to prove that,
\begin{equation}
    \sum_{j=1}^n \sum_{k=1}^{n} a_j a_k F_{RBF}(x_j, x_k) = \sum_{j=1}^n \sum_{k=1}^{n} a_j a_k h(x_j-x_k)
\end{equation}
Thus it is sufficient to prove that, 
\begin{equation}
    \sum_{j=1}^n \sum_{k=1}^{n} a_j a_k h(x_j - x_k) \geq 0
\end{equation}
Therefore, 
\begin{equation}
\begin{split}
    \sum_{j=1}^n \sum_{k=1}^{n} a_j a_k h(x_j - x_k) 
    & = \sum_{j=1}^n \sum_{k=1}^{n} a_j a_k E[e^{i(x_j-x_k)Z}] \\ 
    & = E[\sum_{j=1}^n \sum_{k=1}^{n} a_j e^{i x_j Z} a_k e^{-i x_k Z}] \\
    & = E[| \sum_{j=1}^{n} a_j e^{i x_j Z} |^2] \geq 0,
\end{split}
\end{equation}
This entails that the Radial Basis Function $F_{RBF}$ is positive semi-definite, and thus a kernel.
Without loss of generality, the same proof can be extended when $x$ and $y$ are vectors.
\section{Baselines}
\textbf{Adapted-iCaRL: }
At each incremental step $i$, we first concatenate the exemplar memory and the new class(es) training data, 
\begin{equation}
    \mathcal{D} \leftarrow \mathcal{D}_{exemplar} \bigcup \mathcal{D}_{new}
\end{equation}
where $\mathcal{D}_{exemplar}$ consists of old classes, i.e. $C_0 \bigcup C_1 ... \bigcup C_{i-1}$ and $\mathcal{D}_{new}$ consists of the new classes which the model has to learn, i.e. $C_i$. Thus, $\mathcal{D}$ consists samples from the entire set, $C_0 \bigcup C_1 ... \bigcup C_{i}$

We use the following loss function for the training procedure. 
\begin{equation}
    \mathcal{L} = \alpha * \Sigma_{(x_j,y_j) \in D_{exemplar}} MSE(M_{i-1}(x_j), M_i(x_j)) + (1-\alpha) * \Sigma_{(x_k,y_k) \in D_{new}} MSE(M_i(x_k), y_k) \label{eq:adapted_icarl_loss}
\end{equation}
where $\alpha$ is an hyperparameter in the loss function, we set $\alpha=0.5$ for all the experiments.

We use the herding strategy to sample the exemplar memory, similar to what was used in the original implementation of iCaRL. \\
\\
\textbf{Adapted-EEIL: }
At each incremental step $i$, we first train the model similar to Adapted-iCaRL's training step, which comprises of the data from exemplar memory and the new class(es) data. 
After this step, we perform additional balanced finetuning, as done in EEIL~\cite{castro2018E2EIL}. This balanced finetuning is performed on a training subset containing equal number of samples for each class. 
This is done by sampling $n$ samples for the new class data, $\mathcal{D}_{new}'$, by using the Herding strategy. 
We use the model in the training step performed before balanced finetuning step, $M_i'$, as the teacher network for the Knowledge Distillation loss term. The weights of this model are frozen and it's predicted heatmaps are used for loss calculations. The updated loss term used for this step is given as,  
\begin{equation}
    \mathcal{L}_{old} = \alpha * \Sigma_{(x_j,y_j) \in D_{exemplar}} MSE(M_{i}'(x_j), M_{i}(x_j)) + (1-\alpha) * \Sigma_{(x_j,y_j) \in D_{exemplar}} MSE(M_i(x_j), y_k)
\end{equation}
\begin{equation}    
    \mathcal{L}_{new} = \alpha * \Sigma_{(x_k,y_k) \in D_{new}'} MSE(M_i'(x_k), M_i(x_k)) + (1-\alpha) * \Sigma_{(x_k,y_k) \in D_{new}'} MSE(M_i(x_k), y_k) 
\end{equation}
\begin{equation}
    \mathcal{L} = \mathcal{L}_{old} + \mathcal{L}_{new}
    \label{eq:adapted_eeil_loss}
\end{equation}
where $\alpha$ is an hyperparameter in the loss function, we set $\alpha=0.5$ for all the experiments.

After the balanced finetuning step, we update the exemplar memory by removing samples from the exemplar memory, and adding samples for the new class.

\section{Hyperparameter details}
A list of hyperparameters used in this work is provided in Table~\ref{tab:hyperparameter_1}.
\begin{table}[htp]
\centering
\begin{tabular}{|c|c|}
\hline
\textbf{Hyperparameters}                                                                           & \textbf{Value} \\ \hline \hline
Input Image Size                                                                                       & 512x512                       \\ \hline
Output Heatmap Size                                                                                    & 128x128                       \\ \hline \hline
Base Model Training Epochs                                                                             & 30                            \\ \hline
Base Model Optimizer                                                                                   & Adam                          \\ \hline
Base Model Learning Rate                                                                               & 0.0001                        \\ \hline
Batch size for training Base  Model                                                                    & 13                            \\ \hline \hline
Incremental Model Training Epochs                                                                      & 20                            \\ \hline
Incremental Model Optimizer                                                                            & Adam                          \\ \hline
Incremental Model Learning Rate                                                                        & 0.0001                        \\ \hline
\begin{tabular}[c]{@{}c@{}}Batch Size for training Incremental \\ Model\end{tabular}                   & 5                             \\ \hline \hline
\begin{tabular}[c]{@{}c@{}}Balanced Finetuning Training Epochs \\ (Adapted EEIL)\end{tabular}          & 5                             \\ \hline
\begin{tabular}[c]{@{}c@{}}Balanced Finetuning Optimizer\\ (Adapted EEIL)\end{tabular}                 & Adam                          \\ \hline
\begin{tabular}[c]{@{}c@{}}Balanced Finetuning Learning Rate\\ (Adapted EEIL)\end{tabular}             & 0.00001                       \\ \hline
\begin{tabular}[c]{@{}c@{}}Batch Size for Balanced Finetuning \\ (Adapted EEIL and iCaRL)\end{tabular} & 5                             \\ \hline \hline
$\alpha$ (Adapted EEIL and iCaRL)                                                                      & 0.5                           \\ \hline
\end{tabular}
\caption{Hyperparameter details used in this work.}
\label{tab:hyperparameter_1}
\end{table}
\pagebreak
\section{Additional Experiments}
We perform additional experiments on a different setup, i.e. growing memory case, where the number of samples per class remains fixed. We perform experiments in such a scenario to further demonstrate the efficacy of our approach. We restrict ourselves to 10\% of samples for each class. The results are shown in the Table \ref{tab:growing_memory}. We observe that our proposed DPP w/ clustering performs significantly better than the Random and Herding strategy baselines. Further, our proposed RBF k-DPP ($\gamma = 50$) improves the performance on DPP w/ clustering. 
\begin{table}[htp]
\centering
\begin{tabular}{|c|c|c|}
\hline
\multirow{2}{*}{\textbf{Approach}}                                                                            & \multicolumn{2}{c|}{\textbf{Incremental Steps}} \\ \cline{2-3} 
                                                                                                     & \textbf{1}             & \textbf{2}             \\ \hline \hline
Oracle                                                                                               & 0.8478                 & 0.8457                 \\ \hline
Herding                                                                                              & 0.7513                 & 0.6626                 \\ \hline
Random                                                                                               & 0.7982                 & 0.7259                 \\ \hline
DPP w/ clustering (\textbf{Ours})                                                                             & 0.8245                 & 0.7613                 \\ \hline
\begin{tabular}[c]{@{}c@{}}RBF k-DPP ($\gamma = 50$)\\ (\textbf{Ours})\end{tabular} & \textbf{0.8291}        & \textbf{0.7799}        \\ \hline
\end{tabular}
\caption{PcK@0.05 results for Growing memory (fixed number of samples per class), base classes are \{`cat', `dog', `cow'\} and `horse' and `sheep' are added at the incremental steps. 10\% of \textbf{each} class data is added to the memory. }%We do not remove any samples from the memory.}
\label{tab:growing_memory}
\end{table}

\pagebreak

\begin{figure}[htp]
    \centering
    \includegraphics[width=12cm]{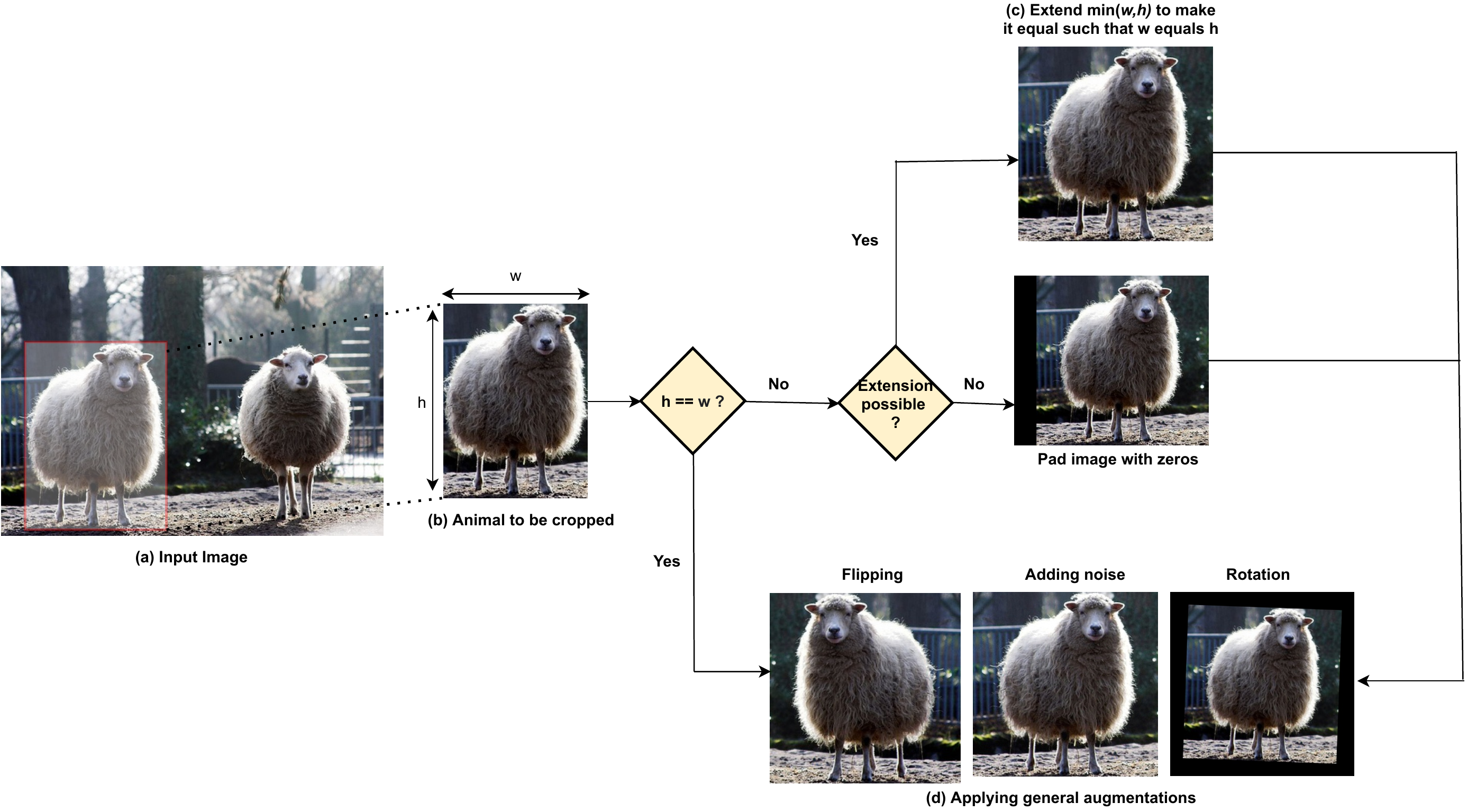}
    \caption{\small{Details of image data pre-processing pipeline. (a) Input Image, (b) we select only the object of interest (animals in our case) as an input to the model, (c) if the selected object of interest has rectangular bounding box we directly extend it or pad zeros to make it a square shaped box and then crop it out, (d) we finally perform various augmentations like flipping, adding noise, rotation and combinations of these.}}
    \label{fig:overview_data_pre_processing}
\end{figure}
\vspace{-0.3in}
\section{Dataset Pre-Processing}
Given an image, which may contain multiple animals in it, we crop out each animal using the ground-truth bounding boxes provided with the dataset. These bounding boxes can be rectangular in shape. In order to explicitly convert them to a square bounding box, we extend the smaller side of the rectangle to make it a square. However there can be an edge case, where while extending the bounding box we may exceed the image boundary region. To overcome this problem, we pad the image with zeros and then extend the smaller side of rectangle to make it a square. After getting a square shaped crop of the animal image, we resize it to a fixed image size across all the input images. We further augment the data by \textit{Flipping, adding Gaussian Noise, Rotating the images by a small random angle}, and a combination of these augmentation strategies. These augmentations helps to increase the training set size and act as a regularizer to reduce overfitting in our pose estimation model. An overview of the data pre-processing pipeline is provided in Figure \ref{fig:overview_data_pre_processing}.
\vspace{-0.1in}
\section{Visualization}
\begin{figure}[htp]
\centering
\begin{tabular}{ccc}
\includegraphics[width=3cm]{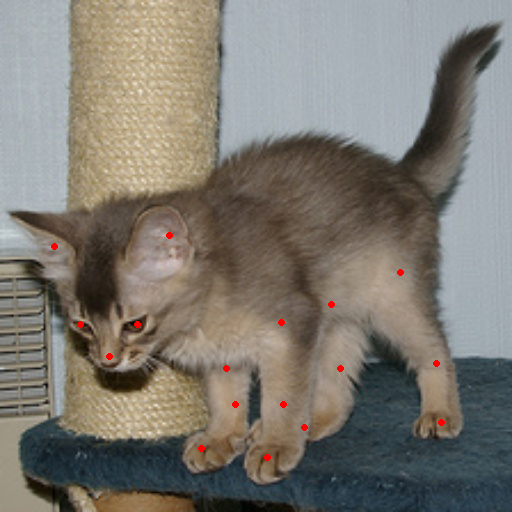}&
\includegraphics[width=3cm]{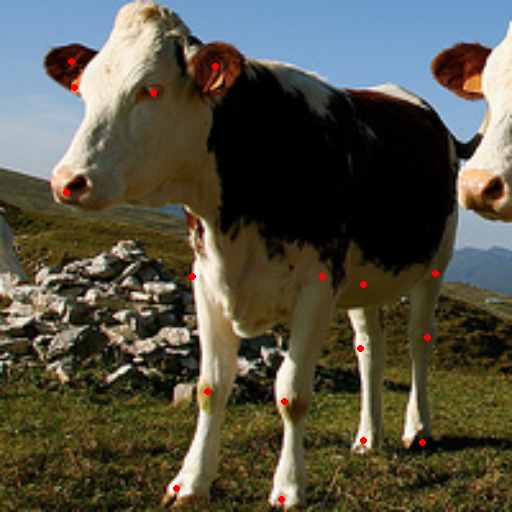}&
\includegraphics[width=3cm]{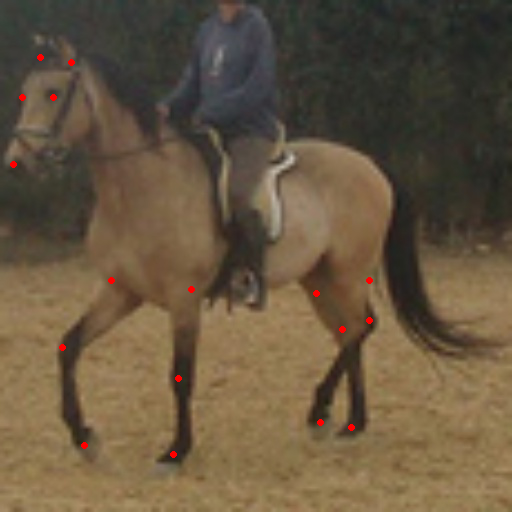}
\\
(a)&(b)&(c)
\end{tabular}
\caption{\small{Visualization of animals and their keypoints labelled.}}
\label{fig:keypoints_labelled}
\end{figure}

We provide visualization of some pre-processed samples and their ground-truth keypoints labelled in Figure \ref{fig:keypoints_labelled}. Red points in the figure show the ground-truth keypoint label. There are total 17 keypoints labelled for each image, namely two `Eyes', two `Earbases', `Nose', four `Elbows', four `Knees' and four `Paws'. 

\begin{figure}[htp]
\centering
\includegraphics[width=0.55\textwidth]{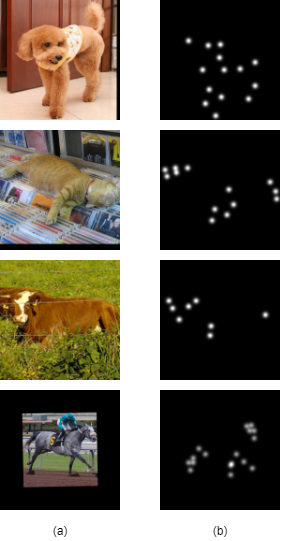}
\caption{The images in (a) are the input images for various classes of animals from Animal-Pose Dataset~\cite{cao2019crossDomainAdaptation}. The images in (b) are \textbf{Summation} of all the 17 heatmaps of the keypoints}
\label{fig:visualization_heatmaps}
\end{figure}

As explained in the main draft, the ground-truth keypoints are converted to Gaussian heatmaps to assist in training the pose estimation model. To generate the Gaussian heatmaps we center each keypoint on the spatial coordinates of the keypoint. A visualization of the input image and the \textbf{summation} of the heatmaps for all the keypoints is provided in the Figure \ref{fig:visualization_heatmaps}.

\bibliography{references}
\end{document}